\documentclass{article}

\PassOptionsToPackage{numbers, compress}{natbib}

\usepackage[final]{neurips_2022}

\usepackage[utf8]{inputenc} 
\usepackage[T1]{fontenc}    
\usepackage{hyperref}       
\usepackage{url}            
\usepackage{booktabs}       
\usepackage{amsfonts}       
\usepackage{nicefrac}       
\usepackage{microtype}      
\usepackage{xcolor}         
\usepackage{subfigure}

\usepackage{graphicx}
\usepackage{amsmath}

\newcommand{\revise}[1]{\textcolor{black}{#1}}

\title{AesopAgent: Agent-driven Evolutionary System on Story-to-Video Production}

\author{
    Jiuniu Wang$^{*}$  \hspace{0.2cm}  Zehua Du$^{*}$  \hspace{0.2cm}  Yuyuan Zhao\thanks{Equal Contribution.}  \hspace{0.2cm}  Bo Yuan \hspace{0.2cm} Kexiang Wang \hspace{0.2cm} Jian Liang \hspace{0.2cm} \\
\textbf{Yaxi Zhao} \hspace{0.2cm} \textbf{Yihen Lu} \hspace{0.2cm} \textbf{Gengliang Li} \hspace{0.2cm} \textbf{Junlong Gao} \hspace{0.2cm} \textbf{Xin Tu} \hspace{0.2cm} \textbf{Zhenyu Guo}\thanks{Corresponding author.} \\ \\
    DAMO Academy, Alibaba Group
}

\begin{document}

\maketitle


\begin{abstract}

The Agent and AIGC (Artificial Intelligence Generated Content) technologies have recently made significant progress. We propose \textbf{Aesop}Agent, an \textbf{A}gent-driven \textbf{E}volutionary \textbf{S}ystem \textbf{o}n Story-to-Video \textbf{P}roduction. AesopAgent is a practical application of agent technology for multimodal content generation. \revise{The system integrates multiple generative capabilities within a unified framework, so that individual users can leverage these modules easily.} This innovative system would convert user story proposals into scripts, images, and audio, and then integrate these multimodal contents into videos. \revise{Additionally, the animating units (e.g., Gen-2 and Sora) could make the videos more infectious.}
The AesopAgent system could \textit{orchestrate task workflow} for video generation, ensuring that the generated video is both rich in content and coherent.  
This system mainly contains two layers, i.e., the Horizontal Layer and the Utility Layer. In the Horizontal Layer, we introduce a novel \textit{RAG-based evolutionary system} that optimizes the whole video generation workflow and the steps within the workflow. It continuously evolves and iteratively optimizes workflow by accumulating expert experience and professional knowledge, including optimizing the LLM prompts and utilities usage. The Utility Layer provides multiple utilities, leading to \textit{consistent image generation} that is visually coherent in terms of composition, characters, and style. Meanwhile, it provides audio and special effects, integrating them into expressive and logically arranged videos. Overall, our AesopAgent achieves state-of-the-art performance compared with many previous works in visual storytelling. \revise{Our AesopAgent is designed for convenient service for individual users}, which is available on the following page: \url{https://aesopai.github.io/}.

\end{abstract}

\section{Introduction}
The recent development of base Large Language Models (LLMs)~\cite{NEURIPS2020_1457c0d6,openai2023gpt4,DBLP:conf/nlpcc/XiaHDZJSCBZ21,DBLP:journals/corr/abs-2303-11381}, and Multimodal Models~\cite{DBLP:journals/corr/abs-2311-04498,DBLP:journals/corr/abs-2304-08485,openai2023gpt4v,DBLP:conf/icml/0008LSH23} has catalyzed significant changes in Artificial Intelligence Generated Content (AIGC). This advancement has led to the effective integration of generative AI technology with traditional Professional Generated Content (PGC) and User Generated Content (UGC), addressing a wide range of user requirements. Notably, technologies such as Stable Diffusion~\cite{9878449}, DALL-E~3~\cite{BetkerImprovingIG}, and ControlNet~\cite{10377881} have excelled in generating and editing high-quality images, attracting significant interests in both academy and industry. Video generation, in contrast to simple static image generation, necessitates managing complex semantic and temporal information, presenting great challenges. Recent initiatives, such as Make-A-Video~\cite{singer2022make-a-video}, Imagen Video~\cite{ho2022imagenvideo}, PikaLab~\cite{pika}, and Gen-2~\cite{10377444}, have demonstrated the ability to produce short videos from textual descriptions. \revise{And Sora~\cite{Sora} is capable of generating high-definition videos up to one minute in length.} However, story-to-video production requires the cooperation of various modules, so gaps remain in image expressiveness, narrative quality, and user engagement. To bridge these gaps, we propose AesopAgent, an agent-driven evolutionary system on story-to-video production. With agent technology, we could effectively harness advanced visual and innovative narrative generation technologies to tackle the complex video generation task, thereby creating videos with compelling narratives and attractive visual effects. \revise{Our system could understand user intentions and identify suitable AI utilities to fulfill user requirements. This feature enables users to effortlessly employ AI capabilities.}

\begin{figure}
    \centering
    \includegraphics[width=1\textwidth]{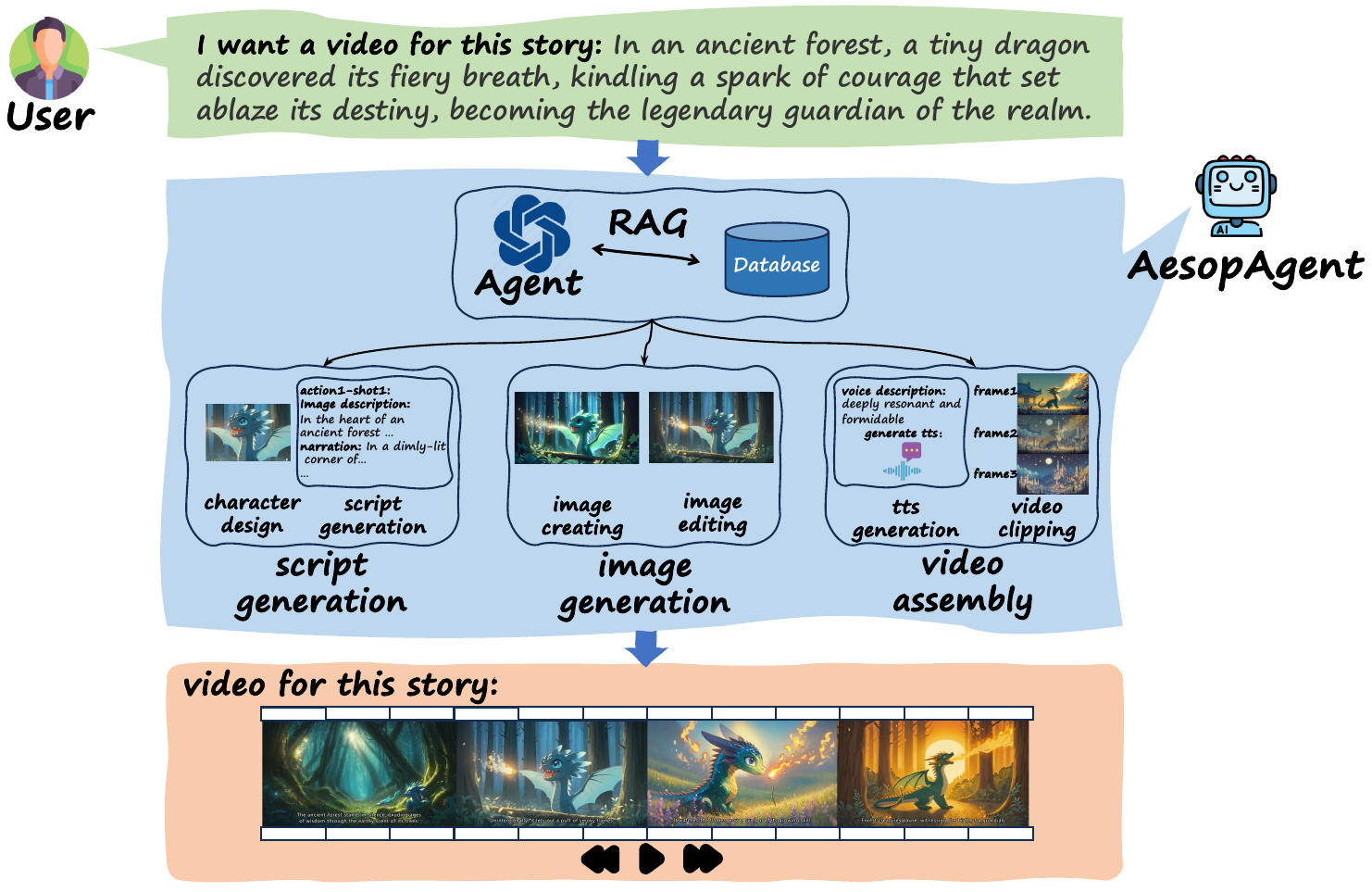}
    \caption{\textbf{Overview of our AesopAgent.} This system would convert the user story proposal into a video assembled with images, audio, narration, and special effects. The video generation workflow suggested by AesopAgent, utilizing agent-based approaches and RAG techniques, encompasses script generation, image generation, and video assembly.}
    \label{fig:teaser_figure}
\end{figure}


AesopAgent utilizes agent-based approaches and RAG techniques, coupled with the incorporation of expert insights, to facilitate an iterative evolutionary process that results in an efficient workflow. This workflow creates high-quality videos from user story proposals automatically. As illustrated in Figure~\ref{fig:teaser_figure}, upon receiving a user's ``Dragon Story'' proposal, AesopAgent employs the well-designed workflow by agents implemented with RAGs, including script generation, image generation, and video assembly, ultimately generates a high-quality dragon story video.

In this paper, our system features a comprehensive process orchestration, including a Horizontal Layer and a Utility Layer. The Horizontal Layer systematically executes high-level strategic management and optimization in the video generation workflow continuously and iteratively. 
The Utility Layer provides a suite of fully functional utilities for each task of the workflow, ensuring the effective execution of the video generation process.
In the Horizontal Layer, we introduce two types of Retrieval-Augmented Generation (RAG) techniques\cite{gao2023retrieval}: Knowledge-RAG and Experience-RAG (denoted as K-RAG and E-RAG). 
E-RAG collects expert experience, and K-RAG leverages existing professional knowledge. Our system constructs and updates the database of E-RAG and K-RAG with the guidance of experts, leading to our evolution ability.
The Utility Layer assembles multiple utilities based on each unit's fundamental capabilities, optimizing their usage based on feedback to ensure stable, rational image composition in terms of character portrayal and style.

The effectiveness of our system is evaluated in three key points: storytelling, image expressiveness, and user engagement. A comparative analysis with systems like ComicAI~\cite{comicai} and Artflow~\cite{artflow} shows that AesopAgent achieves state-of-the-art visual storytelling ability, excelling in coherence, representation, and logic through RAG techniques. Manual evaluations against ComicAI~\cite{comicai} in image element restoration, rationality, and composition indicate AesopAgent's superiority. When compared with parallel video generation research such as NUWA-XL~\cite{DBLP:conf/acl/YinWYWWNYLL0FGW23} and AutoStory~\cite{DBLP:journals/corr/abs-2311-11243}, our system demonstrates leading performance in image complexity and narrative depth. Furthermore, AesopAgent's adaptability is evident in its ability to integrate external software like Runway, catering to specific user needs and underscoring its extensive scalability.

In summary, the contributions of our AesopAgent are threefold: 

1) We have leveraged agent techniques to automate \textit{task workflow orchestration} for video generation, enabling the effective process of converting story proposals into final videos. The workflow is designed by the agent and includes three main steps (script generation, image generation, and video assembly).

2) We proposed a \textit{RAG-based evolutionary system}, collecting expert experience (E-RAG) and professional knowledge (K-RAG) to improve the system performance. The prompts of LLM would be optimized iteratively according to the experience and knowledge in the RAG database, generating narrative coherence text
(e.g., script, image descriptions). 
The RAG techniques also guide the utilities usage, consequently enhancing the visual quality of the generated videos.

3) We achieve \textit{consistent image generation} since our AesopAgent skillfully applies specialized utilities, thus the generated images maintain a cohesive and professional visual expression throughout the shots. The main components of \textit{consistent image generation} are the image composition module, the multiple characters consistency module, and the image style consistency module. There are many utilities in these modules, and some of them have been optimized by our AesopAgent.

\section{Related Works}

\subsection{AI Agent}

Recent advancements in AI agent technology stand at the vanguard of transformative developments within the artificial intelligence domain. A plethora of commendable works involving AI agents have emerged, showcasing groundbreaking progress and innovation. Notable works include: OpenAGI~\cite{ge2023openagi}, AutoGPT~\cite{DBLP:journals/corr/abs-2306-02224}, Voyager~\cite{DBLP:journals/corr/abs-2305-16291}, RET-LLM~\cite{DBLP:journals/corr/abs-2305-14322}, ChatDev~\cite{DBLP:journals/corr/abs-2307-07924}, AutoGen~\cite{DBLP:journals/corr/abs-2308-08155}, etc. These agents exhibit diverse performances in aspects like memory function configuration, task planning capabilities, and modifications to the base Large Language Model (LLM).

As for memory function configuration, recent multi-agent systems (e.g., AutoGen~\cite{DBLP:journals/corr/abs-2308-08155}, MetaGPT~\cite{DBLP:journals/corr/abs-2308-00352}, ChatDev~\cite{DBLP:journals/corr/abs-2307-07924}, AutoGPT~\cite{DBLP:journals/corr/abs-2306-02224})  adopted a shared memory pool for top-level information synchronization, with different technical implementation solutions.  Specifically, MetaGPT~\cite{DBLP:journals/corr/abs-2308-00352} use sub/pub mechanisms, AutoGen~\cite{DBLP:journals/corr/abs-2308-08155} use dialogue delivery. The single-agent systems (e.g., Voyager~\cite{DBLP:journals/corr/abs-2305-16291}, GITM~\cite{DBLP:journals/corr/abs-2305-17144}, ChatDB~\cite{DBLP:journals/corr/abs-2306-03901}) have consensus capabilities (reading, writing, and thinking), combining long-term and short-term memory. In this context, long-term memory structures mostly use historical decision sequences, memory streams, or vector databases, while short-term memory is primarily based on contextual information.

With regard to task planning and utility usage capabilities. The technical solutions for planning capabilities range from prompts and chain mechanisms to standard operating procedures (SOP). Some models, such as Generative Agents and MetaGPT~\cite{DBLP:journals/corr/abs-2308-00352}, enhance execution accuracy and controllability through refined planning and structured output, while models like CAMEL~\cite{DBLP:journals/corr/abs-2303-17760} and ViperGPT~\cite{DBLP:conf/iccv/SurisMV23} focus on using historical information and context to guide the behavior of the Agents. AutoGPT~\cite{DBLP:journals/corr/abs-2306-02224} and AutoGen~\cite{DBLP:journals/corr/abs-2308-08155} models emphasize dialogue and cooperation among multiple Agents to accomplish more complex tasks.

When it comes to modifications of the base model (LLM), The majority of agents (e.g., CAMEL~\cite{DBLP:journals/corr/abs-2303-17760}, ViperGPT~\cite{DBLP:conf/iccv/SurisMV23}, AutoGPT~\cite{DBLP:journals/corr/abs-2306-02224}, ChatDev~\cite{DBLP:journals/corr/abs-2307-07924}, MetaGPT~\cite{DBLP:journals/corr/abs-2308-00352}, AutoGen~\cite{DBLP:journals/corr/abs-2308-08155}) did not modify the basic LLM, indicating a current preference for maintaining its original structure and enhancing performance through peripheral enhancements. While some agents (e.g., ToolLLM~\cite{qin2023toolllm}, OpenAGI~\cite{ge2023openagi}) implemented SFT in the LLM. Particularly for ancillary enhancements, Retrieval-Augmented Generation (RAG) ~\cite{gao2023retrieval, Liu_LlamaIndex_2022} technology has gained increasing traction in fortifying Large Language Models (LLMs). RAG amplifies response accuracy\cite{yu2023chain, yoran2023making} and pertinence by dynamically integrating information from external knowledge bases into LLM responses~\cite{zhang2023retrieve, yu2023augmentation}, effectively mitigating issues such as hallucinations~\cite{huang2023survey}. Its applicability in devising more pragmatic chatbots and applications like QA, summarization, recommendation~\cite{feng2023retrieval, cheng2023lift, rajput2023recommender} has driven rapid technological growth, encompassing methodological and infrastructural domains.


\subsection{Generative Models}

Generative AI models refer to artificial intelligence models capable of autonomously generating new data, such as text, images, video, and audio. They have affected many tasks, including those in natural language processing (NLP), computer vision, and multimodal domains.
In natural language processing, the GPT series, notably GPT-3~\cite{NEURIPS2020_1457c0d6} and GPT-4~\cite{openai2023gpt4}, have revolutionized our comprehension and generation of natural language. These advanced models have given rise to acclaimed AI systems, such as X-GPT~\cite{DBLP:conf/nlpcc/XiaHDZJSCBZ21}, and MM-REACT~\cite{DBLP:journals/corr/abs-2303-11381}. 
In computer vision, generative models are primarily utilized for enhancing image and video generation. Initial approaches often employed techniques like VAE~\cite{DBLP:journals/corr/KingmaW13}, GAN~\cite{Goodfellow2014GenerativeAN}, VQVAE~\cite{DBLP:conf/nips/OordVK17}, or transformers~\cite{DBLP:conf/nips/VaswaniSPUJGKP17}. Diffusioin models (DDPM)~\cite{NEURIPS2020_4c5bcfec} and vision transformer (ViT)~\cite{dosovitskiy2021an} benefit for producing high-quality images and videos. The diffusion models, such as Stable Diffusion~\cite{9878449} and Stable Diffusion XL~\cite{DBLP:journals/corr/abs-2307-01952}, can generate images by denoising certain steps from random noise. Vision transformer (ViT)~\cite{dosovitskiy2021an} based models like CogView2~\cite{ding2022cogview2} and Parti~\cite{DBLP:journals/tmlr/YuXKLBWVKYAHHPLZBW22} can generate image tokens autoregressively. Among these image synthesis models, DALL-E~3~\cite{BetkerImprovingIG} has superior performance, which can generate high-resolution images and excel in converting textual descriptions into vivid images.
There are some recent works~\cite{singer2022make-a-video,ho2022imagenvideo,wang2023modelscope} on video synthesis, and they often employ temporal modules~\cite{singer2022make-a-video} (i.e., temporal convolution and temporal attention) to generate several frames simultaneously.
In the multimodal domain, previous works like NExT-Chat~\cite{DBLP:journals/corr/abs-2311-04498}, LLAVA~\cite{DBLP:journals/corr/abs-2304-08485}, GPT-4V~\cite{openai2023gpt4v}, and BLIP-2~\cite{DBLP:conf/icml/0008LSH23} harmonize language and visual elements, supporting for creating complex and enriched content. Additionally, there are some models (e.g., Audiocraft~\cite{copet2023simple}, MusicLM~\cite{DBLP:journals/corr/abs-2301-11325}, and Music Transformer~\cite{DBLP:conf/iclr/HuangVUSHSDHDE19}) for music generation, demonstrating the adaptability of transformers in handling intricate structures in audio sequences.

\subsection{Story Visualization}

The story visualization task~\cite{DBLP:conf/emnlp/ChenHWNP22, DBLP:conf/eccv/Li22a, DBLP:conf/cvpr/LiGSLCWCCG19, DBLP:conf/emnlp/MaharanaB21} aims to generate visually consistent images or videos from a story described in neural language. Previous story visualization methods mainly focus on conditional text-to-image synthesis~\cite{reed2016generative, Isola2016ImagetoImageTW, zhu2017unpaired}, which can generate high-resolution realistic images. The core of this task is to get the high-quality keyframes conditioned by the story proposals. For example, the input is a complete dialogue session instead of a single sentence in~\cite{sharma2018chatpainter}, and the model is required to generate a visual story. Recently, video generation has been adapted to the story visualization, especially text-to-video~\cite{10.5555/3504035.3504900} or image-to-video generation~\cite{Tulyakov2017MoCoGANDM, Vondrick2016GeneratingVW}. The challenge in video generation for story visualization is to use visual impact to vividly and interestingly describe the given story. DirecT2V~\cite{DBLP:journals/corr/abs-2305-14330} and Phenaki~\cite{DBLP:conf/iclr/VillegasBKM0SCK23} generate different video frames based on various sentences as conditions, forming short video clips. VideoDrafter~\cite{DBLP:journals/corr/abs-2401-01256} and Vlogger~\cite{zhuang2024vlogger} could convert user proposals into scene descriptions, then generate a video clip for each scene. Given key frames of a video, NUWA-XL~\cite{DBLP:conf/acl/YinWYWWNYLL0FGW23} uses the frame interpolating unit to create a video of arbitrary length. Additionally, some systems, like Artflow~\cite{artflow} and ComicAI~\cite{comicai}, could convert user story proposals into videos containing images and audio. However, there is no complete system that can fully display the keyframes, audio, and special effects that are full of story tension in a video. These elements are important for the video to express the story.

\section{Methodology}

\subsection{Overall Framework}

\begin{figure}
    \centering
    \includegraphics[width=1\textwidth]{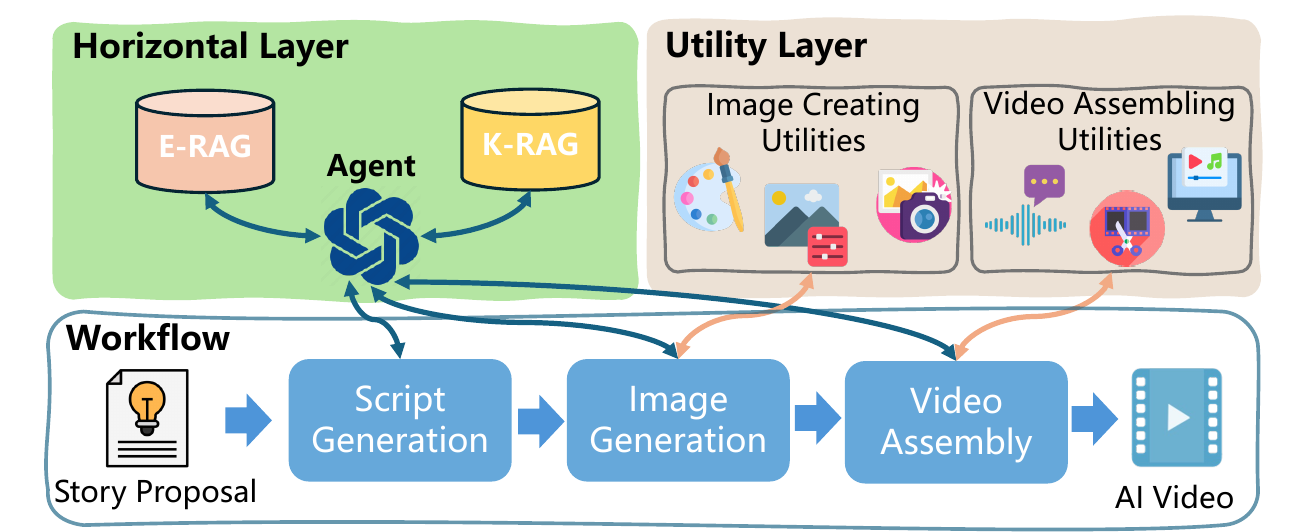}
    \caption{\textbf{Illustration of the AesopAgent framework.} The bottom part of the figure shows the workflow from the user story proposal to the video, and the top part shows the main components of our method: the Horizontal Layer and the Utility Layer. The Horizontal Layer is responsible for leveraging agent and RAG techniques, optimizing workflow and prompts, and optimizing utilities usage, and the Utility Layer is responsible for providing utilities for image generation and video assembly steps.}
    \label{fig:framework}
\end{figure}


The AesopAgent accepts a user story proposal as input and through the implementation of a video generation workflow, it creates a video that conveys the story. As depicted in Figure \ref{fig:framework}, the architecture of the AesopAgent comprises two layers: the Horizontal Layer and the Utility Layer.

The Horizontal Layer employs agent techniques to systematically execute high-level strategic management and optimization in the video generation workflow continuously and iteratively. The expert experience, as well as professional knowledge, are aggregated to construct E-RAG and K-RAG. Then E-RAG and K-RAG are utilized to enhance LLM prompts, so that the Horizontal Layer can employ K-RAG and E-RAG to create and continuously optimize the workflow, further raising requirements for utilities in the Utility Layer and optimizing utilities usage.

The Utility Layer provides a suite of fully functional utilities for each step of the workflow, tailored to task-specific requirements, thus ensuring the effective execution of the video generation process. This layer primarily includes utilities for creating images and utilities for assembling videos. We will introduce the details of the Horizontal Layer and Utility Layer in the following subsections.

\subsection{Horizontal Layer}

In the Horizontal Layer, we leverage agent technology to facilitate workflow creation and optimization with the tasks in the video generation workflow. Our agent $\mathcal{M}$ has the abilities of workflow creating and task planning $\mathcal{M}_T$, prompt execution $\mathcal{M}_e$, and database updating $\mathcal{M}_u$. Our agent combines the accumulated expert experience $E$ and professional knowledge $K$ within the video generation process through interactions with experts. We have developed an efficient and feasible video generation workflow $T$, which is primarily divided into script generation step, image generation step, and video assembly step. 

Moreover, our agents also optimize the prompts to enhance the quality of script generation and also improve utilites usage skills. We propose \textit{RAG-based evolutionary} method (specifically K-RAG and E-RAG) to support these functional parts. K-RAG stores expert knowledge to guide the creation of more professional scripts and utilities usage, while E-RAG summarizes feedback from experts and users, integrating these experiences to refine the video generation workflow. To the best of our knowledge, our AesopAgent is the first system to support task planning and prompt optimization ability based on the RAG techniques.
In the following sections, we will progressively expound on the specifics of each module.

\subsubsection{Task Workflow Orchestration Module}

\begin{figure}
    \centering
    \subfigure[First stage of workflow creation and optimization.]{
    \includegraphics[width=0.456\textwidth]{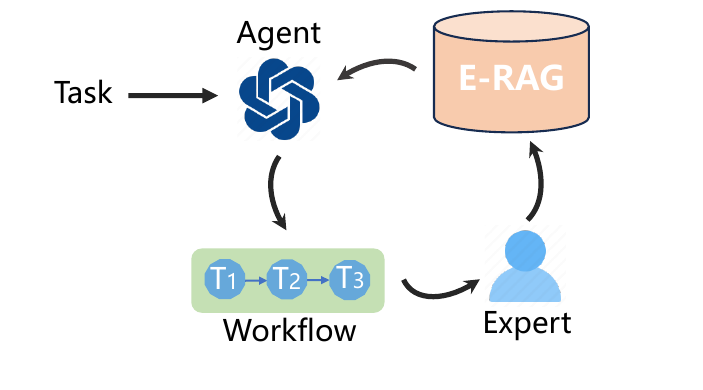}
    }
    \subfigure[Second stage of workflow creation and optimization.]{
    \includegraphics[width=0.511\textwidth]{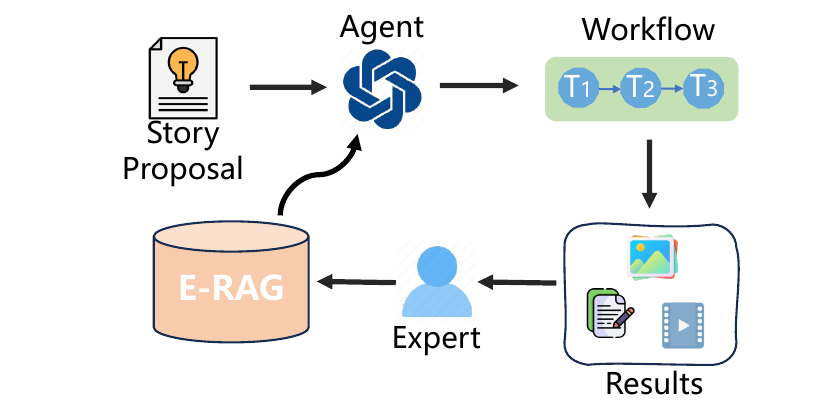}
    }
    \caption{\textbf{The two stages of workflow creation and optimization.} In the first stage, we interact with the agent about a feasible workflow scheme with E-RAG. In the second stage, when we get a feasible workflow scheme, we continue to optimize workflow by combining the results of actual implementation with E-RAG.}
    \label{fig:two_stages}
\end{figure}

The process from a story proposal to a fully realized video is a multifaceted work that encompasses numerous AI methodologies and considerable effort. Establishing an effective task execution workflow is essential for achieving optimal results and enhancing efficiency. Through interaction with agents and leveraging RAG techniques, we collect expert feedback to iteratively optimize the execution workflow, finally, a highly efficient workflow was established.

Our methodology adopts a two-stage strategy for workflow creation and optimization. In the first stage, given a task description $D_T$ and task planning prompt $x_{T
}$, and a task planing agent (LLM) $\mathcal{M}_{T}$, we execute $\mathcal{M}_{T}$ with $D_T$ and $x_{T}$ as inputs, i.e. $\mathcal{M}_{T}(D_T, x_{T})$, to generate an initial workflow $T=(T_{1},\dots, T_{n})$. Subsequently, this workflow undergoes expert evaluation to gauge its practicality, and specific suggestions are provided by experts. E-RAG updates experience to refine the task decomposition process. As the experience of E-RAG accumulates, we can use E-RAG to optimize the workflow. Given experience from RAG: $E=\left \{ e_{1}, \dots, e_{n} \right \}$ and Retriever $\mathcal{R}$, according to the task $T$, retrieve relevant experience $e_{i}$, the agent combines prompt $x_T$ and experience $e_{i}$ to obtain the execution result $r$ from $\mathcal{M}_T(x_T, e_{i})$. In the second stage, once a feasible task segmentation is established, we gather manual feedback from actual task execution results of tasks(such as scripts, images, videos, etc.). E-RAG then utilizes this feedback to further update its experience database and optimize the workflow. The two stages are illustrated in Figure \ref{fig:two_stages}.

\begin{figure}
    \centering
    \subfigure[E-RAG updating mechanism.]{
    \includegraphics[width=0.612\textwidth]{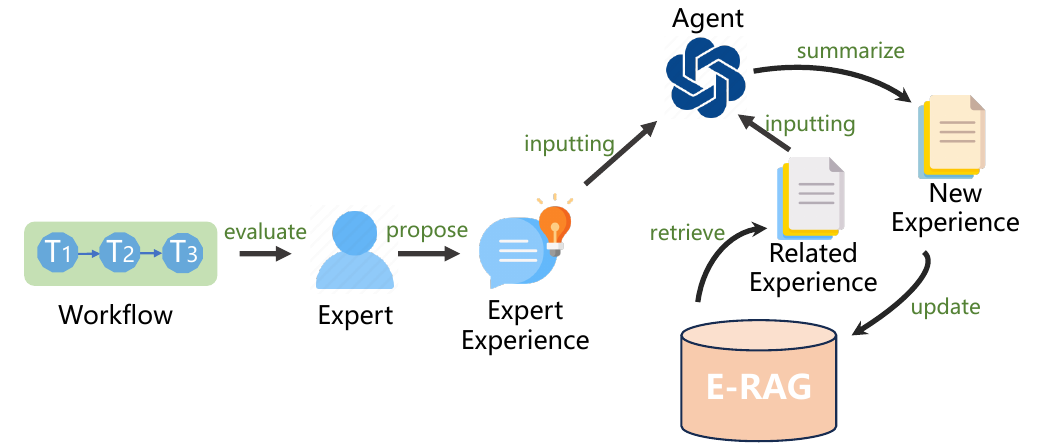}
    \label{fig:E-RAG optimization}
    }
    \subfigure[K-RAG updating mechanism.]{
    \includegraphics[width=0.352\textwidth]{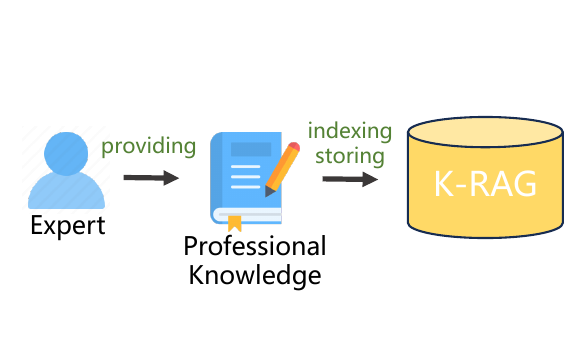}
    \label{fig:K-RAG optimization}
    }
    \caption{\textbf{Two types of RAG updating mechanism.} The left part shows the E-RAG updating process by taking the optimization workflow as an example. The agent summarizes expert experience and relevant experience in the E-RAG dataset, and writes the new experience to the E-RAG database to implement updates. The right part shows the updating process of K-RAG. The K-RAG database is updated by indexing and storing the knowledge documents provided by experts.}
    \label{fig:two RAGs}
\end{figure}

The E-RAG update process is as follows: experts provide feedback based on the result, denoted as $s$. Relying on $s$, the system retrieves the most pertinent experience $e_j$. Subsequently, the agent generates novel experience by synthesizing the feedback and prior experience, represented as $\hat{e}_j$ with agent $\mathcal{M}_u(s, e_j)$, then updating the database of experiences $e_j$ with $\hat{e}_j$ accordingly. If there is no similar experience in the RAG database, the agent formulates a new experience based on the feedback and records it as $e_{n+1}$ using $\mathcal{M}_u(s)$. The E-RAG updating mechanism is illustrated in Figure \ref{fig:E-RAG optimization}.

\begin{figure}
    \centering
    \includegraphics[width=0.88\textwidth]{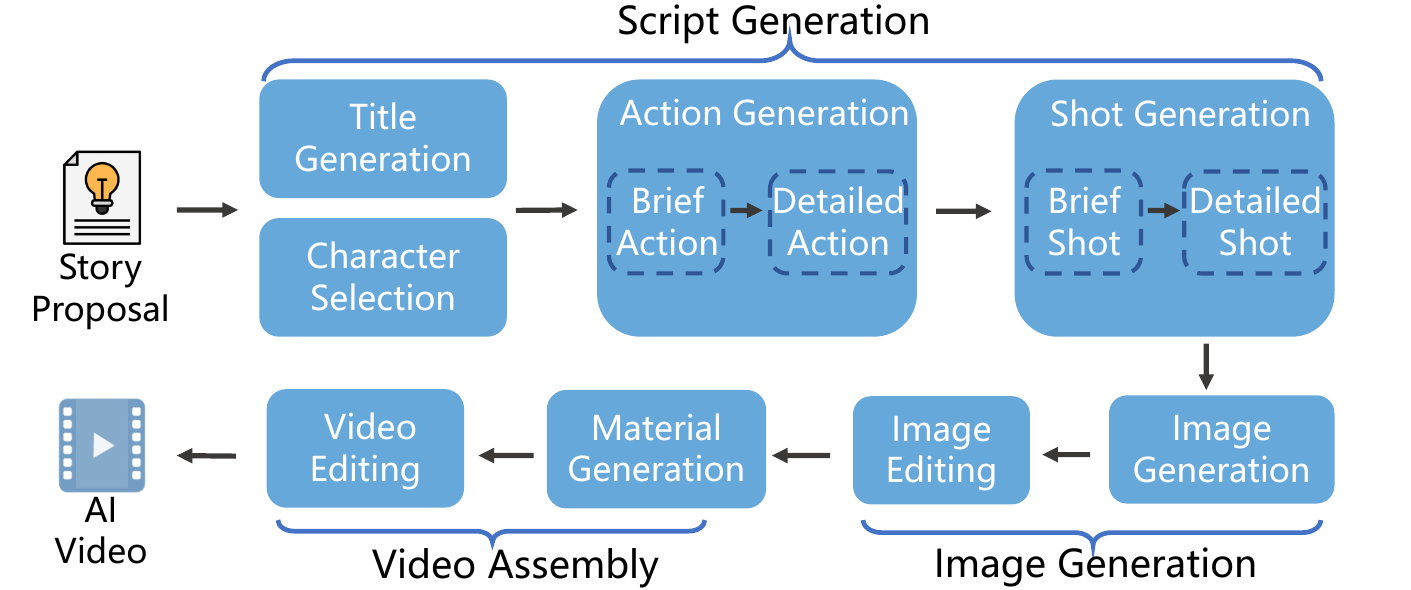}
    \caption{\textbf{The overall workflow of story-to-video production.} The workflow mainly contains three steps: script generation, image generation, and video assembly. The script generation step includes the title generation process and character design and selection process, followed by the action generation and shot generation process. The image generation step mainly includes the image generation process and image editing process. The video assembly step mainly includes the video material generation process and video editing process.}
    \label{fig:workflow}
\end{figure}

Through the iteration of the above processes, we finally get the following workflow, which mainly contains three steps, script generation step, image generation step, and video assembly step. The detailed workflow is in Figure \ref{fig:workflow}. Note that we do not have the optimal workflow, in fact, continuing to run our system and gain experience will make our workflow more reasonable, and our AesopAgent system is constantly improving.

\subsubsection{Prompt Optimization Module}

Our goal is to produce high-quality, captivating storytelling videos. Thus, a professionally crafted script is crucial to achieving the outcomes. In this step, the quality of prompts plays a crucial role. We employ agent technology to iteratively enhance the prompts of the script generation step, culminating in well-structured, narrative-rich scripts that underpin an effective video generation workflow. In this module, we utilize K-RAG to integrate expert knowledge into the prompts of the scriptwriting process and E-RAG to refine prompts based on the feedback from actual results.

\textbf{Providing Professional Knowledge.} The creation of videos with strong storytelling appeal involves numerous professional intricacies, such as shot composition, scene framing, and dialogue arrangement. We deploy K-RAG techniques to offer insights from professional screenwriting and filming practices. The utilize of K-RAG is: given an script prompt $x_s$ and a set of knowledge documents from RAG:$ K=\left \{ k_{1}, \dots, k_{n} \right \}$, the Retriever $\mathcal{R}$, retrieves relevant knowledge $k_{i}$. Subsequently, the agent generates result $r$ from $\mathcal{M}_e(x_s, k_{i})$ based on script prompt $\hat{x}_s$ which is $x_s$ concatenated with domain expertise $k_{i}$. The updating mechanism of K-RAG is depicted in Figure~\ref{fig:K-RAG optimization}, experts provide relevant knowledge documents, and the RAG system performs indexing and storing to write the knowledge into the K-RAG database.

\textbf{Prompt Optimization Based on Feedback.} Large Language Models (LLMs) have significant potential for task execution; however, their performance is heavily reliant on the quality of prompts. In general, the most efficacious prompts are manually crafted through iterative human refinement. In this paper, we propose to use E-RAG to optimize prompts, which can assimilate human experience iteratively. Given a prompt $x$ and a set of experience documents $E=\left \{ e_{1}, e_{2}, ... e_{n} \right \}$ the Retriever $\mathcal{R}$ retrieves pertinent experience $e_{i}$, which facilitates the generation of an optimized script prompt: $\hat{x}_s$ from $\mathcal{M}_e(x_s, e_{i})$. The updating mechanism of E-RAG parallels the mechanism depicted in Figure~\ref{fig:E-RAG optimization}.

With the K-RAG and E-RAG, we iteratively refine the prompts of the script generation effect and ultimately succeed in generating scripts that are both high-quality and engaging for storytelling. As shown in the script generation step in Figure~\ref{fig:workflow}, we could generate the title, select the appropriate characters, and generate information on actions and shots. Here action refers to a shooting scene, generally an action contains several shots.

\subsubsection{Utilities Usage Module}
Given the generated script, the agent system must invoke a suite of utilities to execute the video production, including image generation and video assembly process. Effective usage of these utilities is crucial for achieving high-quality video output.

\textbf{Utility Suggestion \& Creation.} We explore the utilities usage to guide the Utility Layer with K-RAG. We expect that the agent system can understand the provided utilities and actively seek utilities according to requirements. To achieve this,
we first provide some basic utilities and use tutorials according to the workflow, and we write the knowledge into the K-RAG database. When the agent performs a specific task, it retrieves the necessary utilities from K-RAG as required. The agent evaluates whether the current utilities can fulfill the task by comparing the task goals against the utilities usage instructions. If the available utilities are insufficient, the agent could find some related utilities from the web according to the task requirements.

\textbf{Optimization of Utilities Usage.}  
We optimize the utilities usage with E-RAG. Similar to the prompt optimization process, we expect agents to continually optimize their utility usage based on feedback. Taking image utility usage as an example, we use the E-RAG to optimize the image generation workflow. Given image descriptions $d^I$ and experience $\left \{e_{1}, e_{2},... e_{n} \right \}$ from E-RAG database, retriever $\mathcal{R}$, according to the image descriptions $d^I$, the retriever selects the most relevant experience $e_{i}$. The image generation process generates images $\{i^g_1, \dots, i^g_N\}$ using $\mathcal{G}(d^I, e_{i})$, where $N$ is the number of shots for the generated video and $\mathcal{G}(\cdot)$ is the image generation function (containing DALL-E~3~\cite{BetkerImprovingIG} or SDXL~\cite{DBLP:journals/corr/abs-2307-01952}). We evaluate the image against the requirements using manual assessment and multi-modal model(e.g., GPT-4V~\cite{openai2023gpt4v}), modification suggestions $s$ are obtained, which guide the retrieval of experience $e_j$. The agent synthesizes new experience $\hat{e}_j$ from $\mathcal{M}_u(s, e_j)$ using the given suggestions and prior experiences, and subsequently updates the experience in the E-RAG database. Through iteration, we progressively accumulate better image generation experience, thus acquiring more images that align with our objectives.

\subsection{Utility Layer}

\begin{figure}
    \centering
    \includegraphics[width=1.0\textwidth]{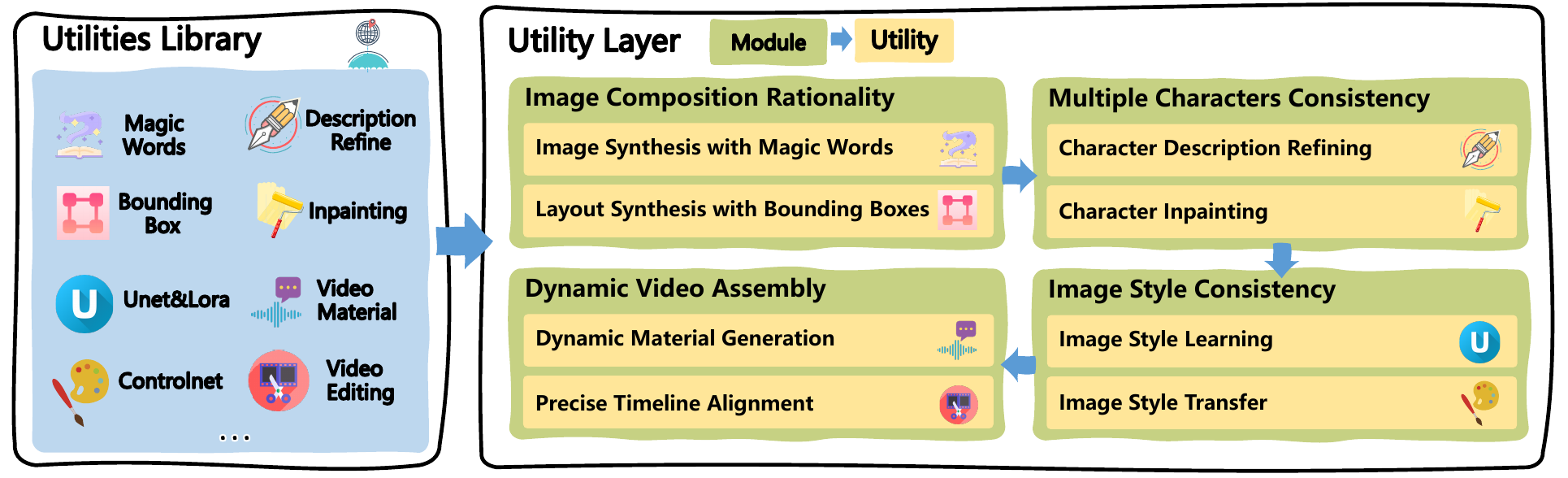}
    \caption{\textbf{The illustration of Utility Layer.} The Utility Layer contains four modules, i.e., image composition rationality, multiple characters consistency, image style consistency, and dynamic video assembly. Each module has some utilities from the utilities library, and these utilities are created and optimized by the Horizontal Layer.}
    \label{fig:tool}
\end{figure}

Given the user's story proposal, our system designs a script of $N$ shots with image descriptions $d^I = \{d^I_1, \dots, d^I_N\}$. We aim to generate images $\{i^g_1, \dots, i^g_N\}$ that match the image descriptions, then assemble video material (i.e., images, audio, and narrates) into a dynamic video to show a complete story. 
As shown in Figure~\ref{fig:tool}, the Utility layer contains four modules, i.e., image composition module, multiple characters module, image style module, and video assemble module. We propose an innovative approach to integrate utilities into these four modules. The former three modules contribute to our \textit{consistent image generation}.

\subsubsection{Image Composition Rationality}

The image composition module aims to generate images $\{i^{co}_1, \dots, i^{co}_N\}$ with composition rationality. 
This module is implemented with the following two utilities, i.e., image synthesis with magic words and layout synthesis with bounding boxes.

\textbf{Image Synthesis with Magic Words.} We employ an image synthesis unit to convert the image description $d^I_t$ to a well-composed image $i^{co}_t$ for the $t$-th shot of the video. During the training process of the image synthesis unit, certain ``magic words'' have been annotated within the training dataset, enabling the image synthesis unit to comprehend their meanings. Then these composition-related magic words are used as prefixes to generate images that meet specific compositional requirements. Typical magic words include ``Middle view'', ``Close view'', ``Low Angle'', ``High Angle'', etc.

\textbf{Layout Synthesis with Bounding Boxes.} To fulfill more detailed compositional demands, the $t$-th image $i^{co}_t$ could also employ bounding boxes $\{box_{1}^t, \dots, box_{k_b}^t\}$ as another controllable condition, where $k_b$ is the number of bounding boxes for the $t$-th image. This allows for generating objects in fixed locations corresponding to the given descriptions. This approach, known as Layout Diffusion~\cite{DBLP:conf/cvpr/ZhengZLQSL23}, can put different characters into specified locations to create a cohesive scene within one image.

\subsubsection{Multiple Characters Consistency}

The multiple characters consistency module aims to ensure the consistent characters appearance for the images of the same story. 
We provide two utilities in this module, i.e., character description refining and character inpainting.
So that the multiple characters module turns the well-composed images $\{i^{co}_1, \dots, i^{co}_N\}$ into the characters consistant images $\{i^{ch}_1, \dots, i^{ch}_N\}$. 

\textbf{Character Description Refining.} Our approach refines two types of character descriptions for consistency: attached and separate character descriptions. The attached character description, embedded within the overall image description, encompasses key characteristics such as gender, age, clothing, and hairstyle. This integration ensures the generation of similar characters in different images based on the attached character description within the $t$-th shot's image description $d^I_t$. Additionally, to portray characters more specifically, we generate separate character descriptions $d^{ch}_t = \{d^{ch}_{t,1},\dots,d^{ch}_{t,k_c}\}$ for each main character in the story as the semantic condition for Character Inpainting, where $k_c$ is the number of main characters in the $t$-th shot.

\textbf{Character Inpainting.} We segment the $i^{co}_t$, and obtain the region mask $m^{ch}_t = \{m^{ch}_{t,1},\dots,m^{ch}_{t,k_c}\}$ for $k_c$ characters. Subsequently, the character inpainting is performed on the image at each position of $m^{ch}_t$ to obtain an image $i^{ch}_t$ that better aligns with the separate character description $d^{ch}_t$. This process can be expressed as 
\begin{align}
 i^{ch}_t = \text{Inp}(i^{co}_t, m^{ch}_t, d^{ch}_t)\,,
\end{align}
 where $\text{Inp}(\cdot)$ represents the character inpainting unit, re-illustrating the character at $m^{ch}_t$ based on the separate character description $d^{ch}_t$. The diffusion model is the core of the character inpainting unit, and we employ technology such as DreamBooth~\cite{DBLP:conf/cvpr/RuizLJPRA23} and the parameter fusion of multiple models to gain consistent character generation.

\subsubsection{Image Style Consistency}

The multiple characters module generate character consistent images $\{i^{ch}_1, \dots, i^{ch}_N\}$. The image style module makes these images uniform in style, gaining the final generated images $\{i^{g}_1, \dots, i^{g}_N\}$. 
We choose two utilities to support image style consistency, i.e., image style learning and image style transfer.

\textbf{Image Style Learning.} To enable the image synthesis unit to generate images in a specific style $s_{ty}$ (e.g., colorful comic style), we first collect a set of images in that style for training, allowing the unit to incorporate the style into its parameters of deep neural networks. In detail, these parameters could be the UNet~\cite{DBLP:conf/miccai/RonnebergerFB15} or LoRA~\cite{DBLP:conf/iclr/HuSWALWWC22} layers. The image synthesis unit is trained with the collected image-text pairs. Once trained with images of a fixed style, the unit can consistently generate images in a certain style.

\textbf{Image Style Transfer.} Here the ControlNet~\cite{10377881} unit is employed for image-style transfer 
, so that the generated images could have the same style.
We also obtain the depth map $m^{dp}_t$ of image $i^{ch}_t$ as a condition. So the ControlNet~\cite{10377881} unit $\text{CT}(\cdot)$ could be employed to transform the image $i^{ch}_t$ and image description $d^I_t$ into $i^g_t$ as,

            \begin{align}
i^g_t = \text{CT}(s_{ty}, d^I_t, i^{ch}_t, m^{dp}_t, \lambda_{ct}) \,,
\end{align}
where the hyperparameter $\lambda_{ct}$ aims to balance the editing intensity, controlling the weights of different channels within ControlNet~\cite{10377881}.

\subsubsection{Dynamic Video Assembly}

The dynamic video assembly module could be split into two utilities, i.e., dynamic material generation and precise timeline alignment.
Unlike traditional simple editing rules and template-based approaches, this module integrates video content with audio and transition effects. 

\textbf{Dynamic Material Generation.} Dynamic material generation automatically selects and edits materials based on the narrative requirements, focusing on audio and video material generation. Our system incorporates background music, sound effects, and voice elements for audio, with the voice component generated using Text-to-Speech (TTS) technology. The goal of video material editing is to transform storyboard images into dynamic videos. This includes detailed descriptions and visual effect analyses (e.g., push and pull shots, rotations, zooms) and various transition effects (e.g., dissolves, wipes, and pushes).

\textbf{Precise Timeline Alignment.} Timeline alignment assembles different materials into a cohesive video. This utility ensures that all selected and edited materials are synchronized on the main timeline. The system employs precise timing adjustments to maintain video coherence, addressing the original material temporal misalignment during camera movement and transitions. These features simplify the editing process, enhancing system efficiency and video coherence.

\section{Experiments}

\subsection{Main Results of Horizontal Layer}

\subsubsection{Task Workflow Orchestration}
In this section, we discuss the enhancements that agent technology contributes to task workflow orchestration and optimization. We show this process through a case study.

\begin{figure}
    \centering
    \includegraphics[width=1\textwidth]{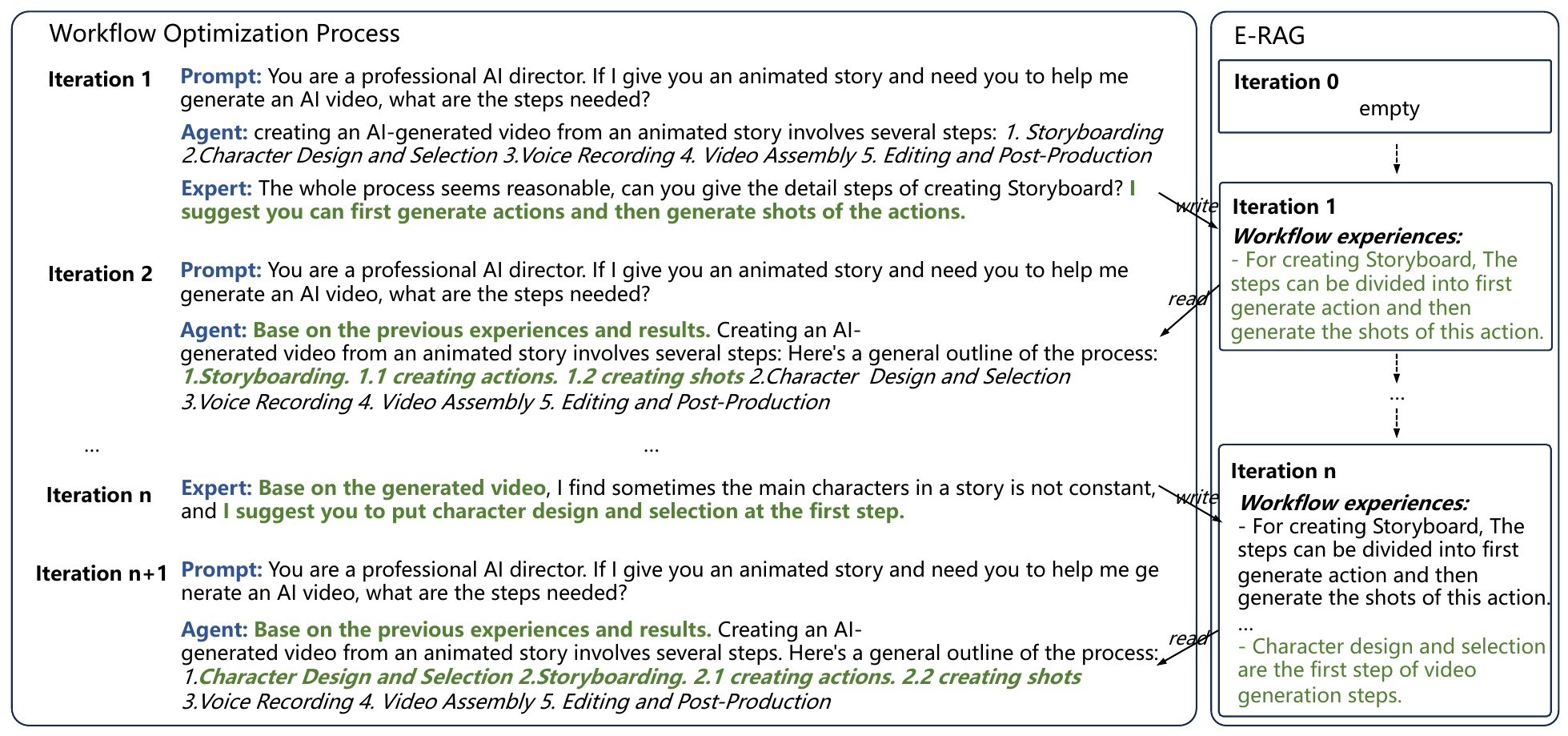}
    \caption{\textbf{Case of workflow optimization using E-RAG.} The text in green is the result of each step's update, we note that by using E-RAG, the agent system has learned from experts' experience for optimizing workflow.}
    \label{fig:workflow_optimization}
\end{figure}

\textbf{Workflow Optimization.} We use the E-RAG technique to iteratively optimize the process of workflow, as depicted in Figure \ref{fig:workflow_optimization}. We illustrate our technology by presenting typical cases in the optimization process. At the beginning of the iteration, the E-RAG database is empty, and experts give suggestions for decomposing the storyboard into actions and shots according to the previous result. These suggestions are summarized by the agent and written into our E-RAG database. In the subsequent iteration, the agent leverages its accrued experience and previous results to propose a more reasonable workflow. After $n$ iterations, we observed that the main characters in different actions of the produced videos lacked stability, so we proposed to put character design and selection as the first step of the workflow to ensure the stability of the main characters throughout the story. Similarly, by utilizing E-RAG techniques, we successfully optimized the workflow.

\subsubsection{Prompt Optimization and Script Evaluation}

In this section, we show how RAG improves prompt optimization. We show the optimization process of prompt in the script generation step through a case study. We invite professional screenwriters to evaluate the scripts generated by our method to show the comparison between our professional knowledge of prompt optimization. We also show the comparison with other systems from the evaluation scores.

\begin{figure}
    \centering
    \includegraphics[width=1\textwidth]{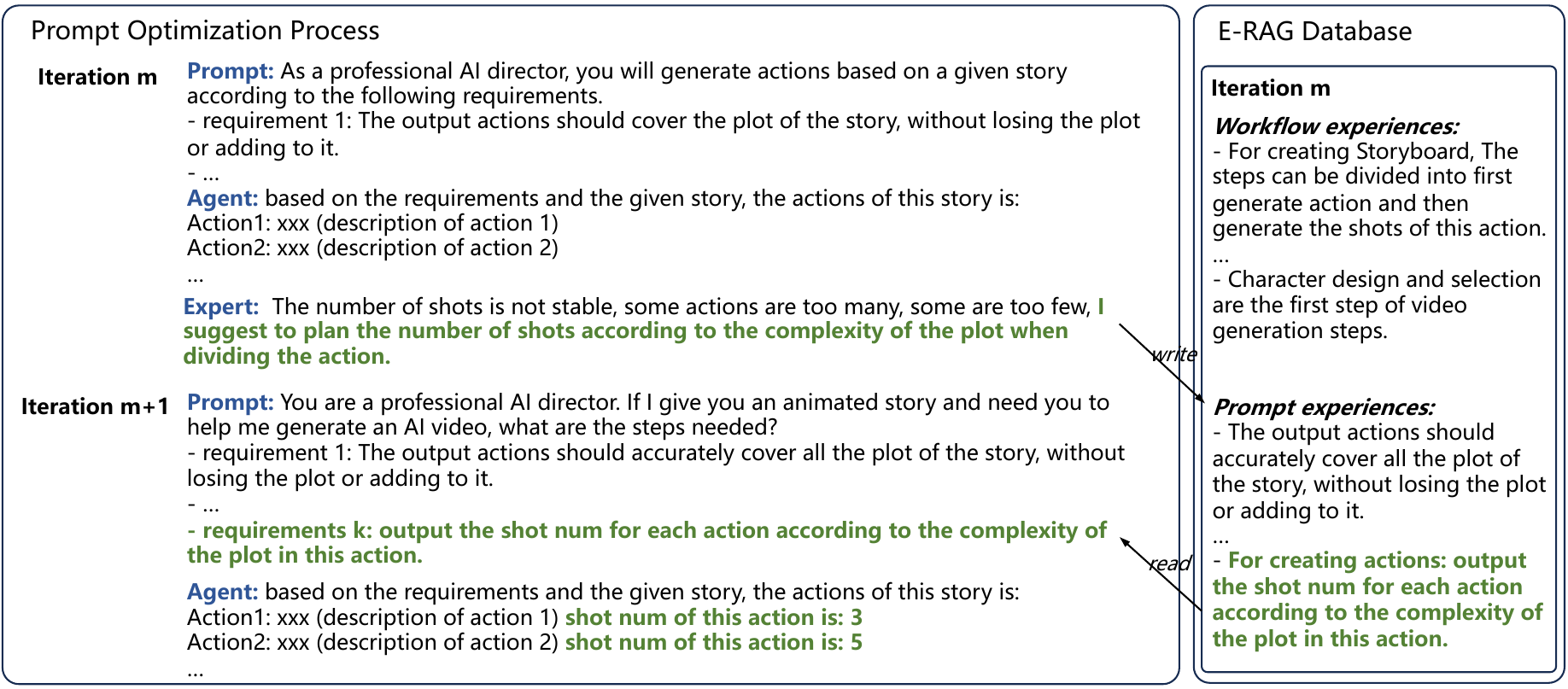}
    \caption{\textbf{Case of prompt optimization using E-RAG.} The text in green is the result of each step's update, we note that by using E-RAG, the agent system has learned from experts' experience for optimizing prompts.}
    \label{fig:prompt_optimization}
\end{figure}

\textbf{Prompt Optimization.} In a similar vein, agent technology can be applied to the enhancement of prompt optimization, as illustrated in Figure~\ref{fig:prompt_optimization}. The expert observed that the allocation of shot numbers for each action did not correspond to the complexity of the story, thereby it is suggested to add global shot number planning during the action division step, which will make the generated shots more reasonable. Similarly, in the application of E-RAG techniques, we successfully implemented such enhancements in prompt optimization.

\textbf{Script Evaluation.} We engaged professional scriptwriters to evaluate our scripts. They established evaluation criteria for \textbf{completeness} (evaluate the completeness of the story structure, the completeness of the content in each shot etc.), \textbf{fidelity} (evaluate the degree of reduction of the story etc.), and \textbf{logical coherence} (evaluate whether the story plot and the arrangement of the script is logical coherence etc.) to conduct a comprehensive assessment of the script quality. We calculated an overall score based on the weighted sum of these three scores to gauge the overall quality of the script. We compared the results of our method with those obtained without professional knowledge (K-RAG), as well as with the best story creation applications currently available on the market: Artflow~\cite{artflow} and ComicAI~\cite{comicai}. Below are detailed analyses:

\begin{table}[]
\centering
\scalebox{0.77}{
\begin{tabular}{l|ccc|c}
\hline\hline
                         \textbf{Methods}       & \textbf{Completeness (30\%)} & \textbf{Fidelity (30\%)} & \textbf{Logical Coherence (40\%)} & \textbf{Overall} \\ \hline
ComicAI~\cite{comicai}                         &  85  &  33  &  78  &  66.6  \\ 
Artflow~\cite{artflow}                      &  \textbf{98} &  43  &  87  &  77.1  \\ 
AesopAgent w/o professional knowledge &  92  &  32  &  85  &  71.2  \\ 
AesopAgent (Ours)                &  94  &  \textbf{60}  &  \textbf{89}  &  \textbf{81.8}  \\ \hline\hline
\end{tabular}}
\caption{\textbf{Script evaluation results.} The best results are shown in bold.}
\label{table:script_evluation}
\end{table}

\textit{Comparison of Professional Knowledge.}
We first compared the scripts generated by our method with our method without expert knowledge (where K-RAG was not used during the script generation step). The absence of expert knowledge during the script generation process means there is a lack of guidance for crafting the scripts, which can lead to some unprofessional aspects such as scene arrangement and plot segmentation, resulting in poorer outcomes, as shown by the results in Table~\ref{table:script_evluation}, there was a decline in completeness, fidelity, and logical coherence, particularly in fidelity. This decrease can be attributed to the lack of expert guidance, which makes the language model generate freely, leading to story error propagation and significantly reduced fidelity. This demonstrates the importance of professional knowledge (K-RAG) for the quality of script generation.

\textit{Comparison with Other Methods.}
We also compared our method with two of the best-performing story creation applications currently on the market: Artflow~\cite{artflow} and ComicAI~\cite{comicai}, with results presented in Table \ref{table:script_evluation}. The results indicate that our method performed well on all these dimensions. Our scripts maintained the highest level of fidelity and logical coherence, even while our method allowing for the appropriate expansion of simple stories. This indicates that our script achieved a good narrative effect. For completeness, Artflow~\cite{artflow} and our AesopAgent are both good, but as we need to consider the final effect of the video, for generating narration, we require that the content shown in the image cannot be repeatedly described in narration, so we will lose some points when we do the script evaluation (only consider the text part), this is as expected. Overall our AesopAgent showed a significant improvement in the overall score and effect compared to other methods at the script generation stage. Professional screenwriters gave a high evaluation of our overall scripts and the final video.

\subsubsection{Utilities Optimization}

In this section, we will take the use of image generation utilities as examples to demonstrate the effect of our agent technology on the design of utilities and the optimization of utilities usage.

\textbf{Utility Suggestion \& Creation.}
\begin{figure}
    \centering
    \includegraphics[width=1\textwidth]{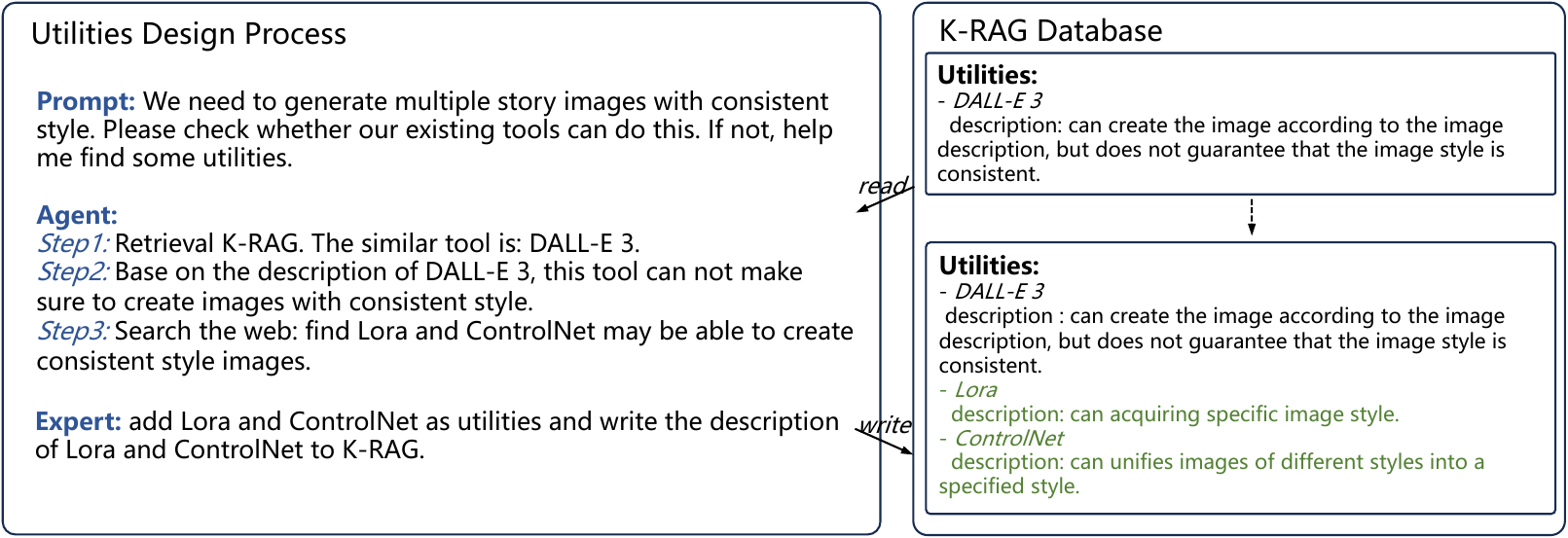}
    \caption{\textbf{Case of Utility Suggestion \& Creation with K-RAG.} The text in green is the result of each step's update, we combined K-RAG and agent technology to help us complete the utility design of style-consistent images.}
    \label{fig:utilities_proposal}
\end{figure}

\begin{figure}
    \centering
    \includegraphics[width=0.81\textwidth]{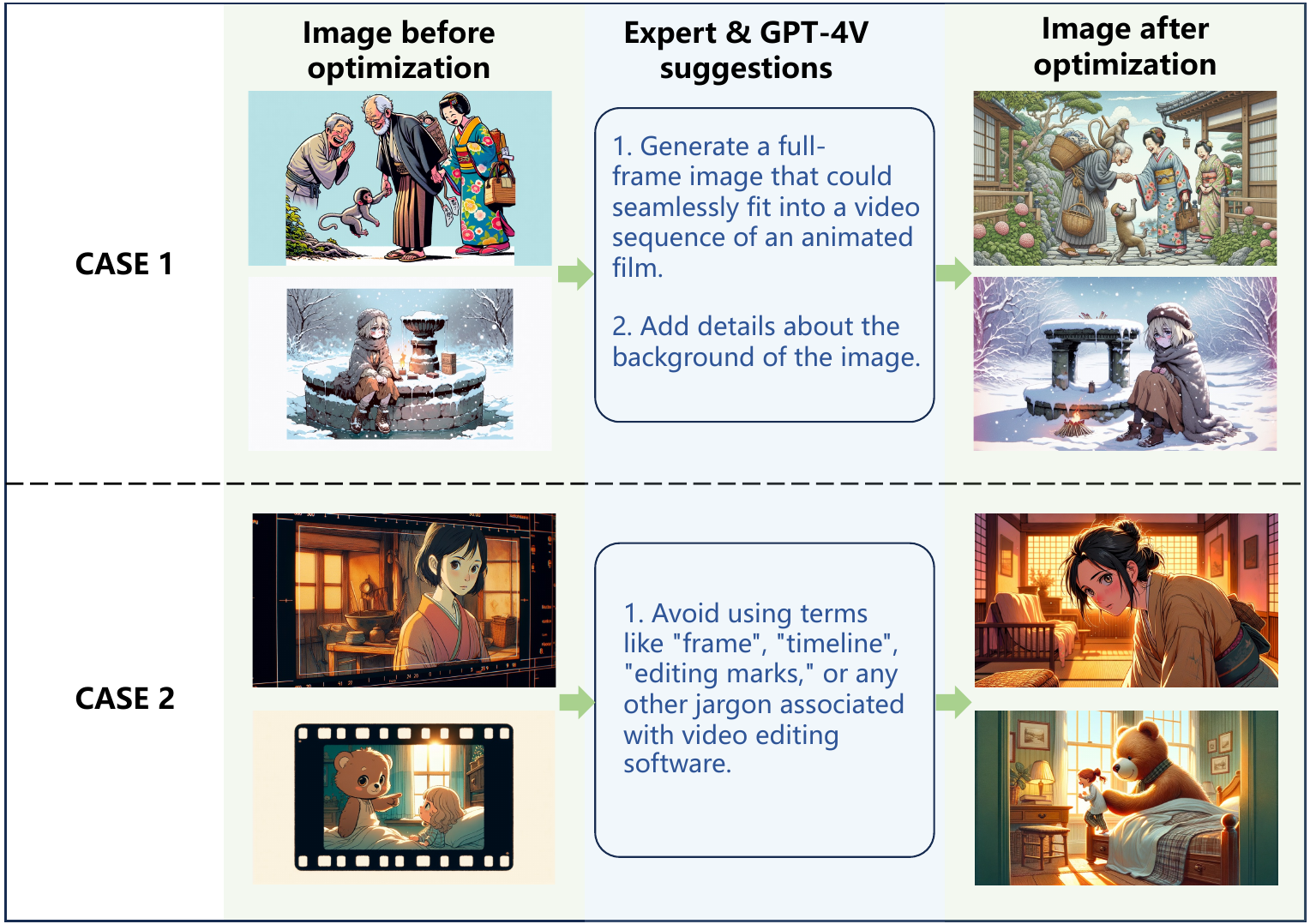}
    \caption{\textbf{Cases of utility usage optimization using E-RAG.} We present two cases of the use of E-RAG learning experts and GPT-4V experience to improve the quality of image generation.}
    \label{fig:utilities_usage}
\end{figure}
Leveraging agent technology has enabled us to effectively select and employ utilities that meet specific design requirements. We take the need for style consistency in the images of our stories as a case. As depicted in Figure~\ref{fig:utilities_proposal}, upon consulting our agent about the availability of utilities within the K-RAG framework that could satisfy our requirements, the agent analyzed and identified DALL-E~3 as the closest match. However, DALL-E~3 was found incapable of ensuring style consistency. As a result, the agent conducted an online search and proposed a solution combining Lora and ControlNet that could potentially achieve the desired style uniformity. Following a thorough evaluation by our experts, this proposed method was confirmed to be effective, culminating in the incorporation of both Lora and ControlNet into the K-RAG database.

\textbf{Utilities Usage Optimization.} Similar to the optimization process for prompts described in Figure~\ref{fig:prompt_optimization}, we utilized E-RAG techniques to refine the usage of our image generation utilities. To generate images that align with the requirements of our narratives, we imposed certain constraints on the usage of these utilities. In practice, we sometimes produced images that did not fulfill the video criteria. Leveraging the insights of domain experts and the feedback from GPT-4V, we refined the image descriptions, which significantly decreased the likelihood of such discrepancies. As depicted in Figure~\ref{fig:utilities_usage}, in case 1, we occasionally produced images lacking backgrounds or with inappropriate framing. Experts and GPT-4V suggest adding more detailed background descriptions and incorporating framing restrictions, the quality of the resulting images was noticeably enhanced. Case 2 illustrates the generation of images resembling film strips or appearing as if they were in the editing phase. By avoiding certain terminologies as suggested by experts and GPT-4V, the probability of generating this type of image is greatly reduced.

\subsection{Main Results of Utility Layer}

Here we show the results of the Utility Layer, including the case analysis and user study. We compare the generated images of our AesopAgent with other methods (i.e., SDXL~\cite{DBLP:journals/corr/abs-2307-01952} and ComicAI~\cite{comicai}) both qualitative and quantitative to show the effectiveness of the Utility Layer.


\begin{table}[]
\small
\centering
\begin{tabular}{l|p{12cm}}
\hline  \hline
\textbf{Index} & \textbf{Image Description} \\ \hline \hline
   \textbf{Image 1}     &    A young girl with long golden curls on her face, her eyes wide and full of wonder. Wearing a simple blue and white ruffled dress, she has a mischievous look in her twinkling blue eyes, accentuated by her small, upturned nose. Little girl standing in front of an antique wooden house.  \\ \hline
   \textbf{Image 2}     &    In the heart of a quaint kitchen filled with rustic wooden furniture and the aroma of breakfast, a young girl with long, curly golden hair and twinkling blue eyes is captured in a moment of anticipation. Her attire is a soft blue and white dress, ruffled from play. The girl holds a spoon precariously in her petite hand, raising it to her mouth as she gently blows on the steaming porridge scooped from the smallest bowl on the table.     \\ \hline
   \textbf{Image 3}     &      Before an arrangement of three different-sized beds—each described as grand, moderate, and petite—stands a young girl with long, curly golden hair that cascades down her back, her shoulders drooped in exhaustion. In front of the broad first bed with a firm mattress, the girl presses her hands down on the bedding, her expression weary with a slight frown and a glint of hope in her twinkling blue eyes.    \\ \hline
   \textbf{Image 4}     &     In a cozy, warm-toned kitchen, a young boy, possibly six or seven years old with short, curly, and unkempt hair and dressed in a slightly oversized soft shade of blue shirt and shorts that hit just above the knees, exhibits a kind, innocent expression. Eyes sparkling with wonder, the boy reaches out with his small hands, exploring a golden cake with gentle curiosity. Presenting the cake is a middle-aged woman with a benevolent smile, her hands showing signs of hard work, and wearing a dress with a full-length apron splattered with flour.    \\ \hline
   \textbf{Image 5}     &      Morning sunlight streams through a window into a quaint rustic kitchen, illuminating the earthenware pots and fragrant herbs lining the shelves with a warm, amber glow. A young boy with short, curly hair and a kind expression stands at the center. The boy is about six or seven years old, dressed in a simple, oversized blue shirt and shorts, with well-worn shoes. A straw hat, with leaves peeking out, sits precariously atop his head, and a pound of butter, wrapped in leaves, is visible underneath the hat. A middle-aged woman with fine lines on her face, her hair held back by a cloth headband, stands to the side, her face showing a blend of affection and apprehension.    \\ \hline
   \textbf{Image 6}     &        Under the relentless sun reigning high in the azure expanse, light pours radiance over the countryside. In the sweltering heat, a young boy around six or seven years old, with an innocent expression, short curly hair, dressed in a slightly oversized soft blue shirt, shorts just above the knees, and well-worn shoes, makes his way across the landscape. The golden cake that once resided in the young boy's hand now crumbles amid the overpowering warmth.    \\ \hline \hline
\end{tabular}
\caption{\textbf{The selected image descriptions for generated image comparison.} We select six image descriptions from two stories. The Image 1 to 3 are from the story ``Goldilocks'' and the Image 4 to 6 are from the story ``Epaminondas and Auntie''.}
\label{tab:case_image_description}
\end{table}

\subsubsection{Effectiveness of Utility Modules}

\begin{figure}
    \centering
    \includegraphics[width=0.99\textwidth]{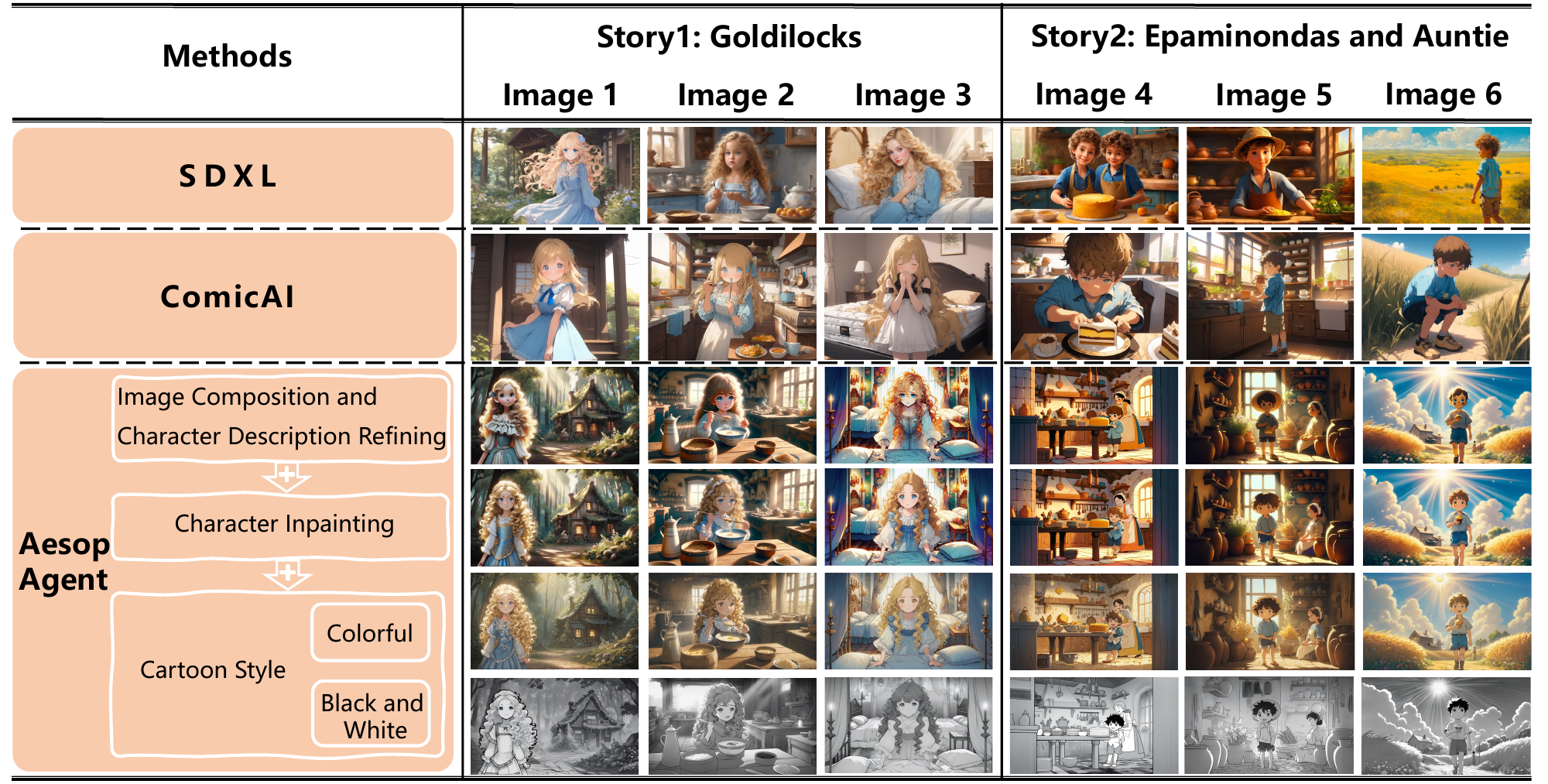}
    \caption{\textbf{Qualitative results of different methods on image generation.} We show the generated images from two stories (i.e., ``Goldilocks'' and ``Epaminondas and Auntie'') from SDXL~\cite{DBLP:journals/corr/abs-2307-01952}, ComicAI~\cite{comicai}, and our AesopAgent. Specifically, the generated images through different AesopAgent's utility modules are shown in Columns 3 to 6.}
    \label{fig:utility_modules}
\end{figure}

This experiment demonstrates the performance of our utilities. As shown in Table~\ref{tab:case_image_description}, we select image descriptions from two stories (i.e., ``Goldilocks'' and ``Epaminondas and Auntie'') to explore the generated images from different methods. The generated images are displayed in Figure~\ref{fig:utility_modules} where SDXL~\cite{DBLP:journals/corr/abs-2307-01952} and ComicAI~\cite{comicai} are employed as baselines. Here ``Goldilocks'' shows the blondie in various backgrounds, while ``Epaminondas and Auntie'' illustrates the interactions between two characters (i.e., a boy and his auntie). 

Our utilities include three modules specifically, i.e., the image composition and character description refining module, the character inpainting module, and the image style consistency module. The image composition and character description refining module manages scene layout and character attributes. As indicated in Image 4, the interaction between the ``Epaminondas'' and ``Auntie'' is depicted accurately and vividly according to the given image description. The Character Inpainting module enhances character portrayal. For example, the ``Goldilocks'' in Image 1 and the ``Epaminondas'' in Image 5 are inpainted to change their skin tones to keep their image consistent. The image style consistency module effectively rendered the overall scene's aesthetic.
Constrasted to SDXL~\cite{DBLP:journals/corr/abs-2307-01952} and ComicAI~\cite{comicai}, our utilities exhibit enhanced control in style and multi-character consistency. For instance, in Image 4, SDXL~\cite{DBLP:journals/corr/abs-2307-01952} draw the character ``Auntie'' as a boy, and ComicAI~\cite{comicai} ignores the character ``Auntie'', while our utilities can control the consistency of the characters (i.e., ``Epaminondas'' and ``Auntie'') and reserve the interaction among them. With the image style consistency module, we can render the images in different styles as shown in Figure~\ref{fig:utility_modules}.  ComicAI~\cite{comicai} could only use colorful cartoon style, and SDXL~\cite{DBLP:journals/corr/abs-2307-01952} does not focus on any certain style. 
Consequently, our Utility Layer achieves the industry standard, ensuring compositional coherence, consistent character depiction, and stylistic uniformity in image generation.

\begin{figure}
    \centering
    \includegraphics[width=1\textwidth]{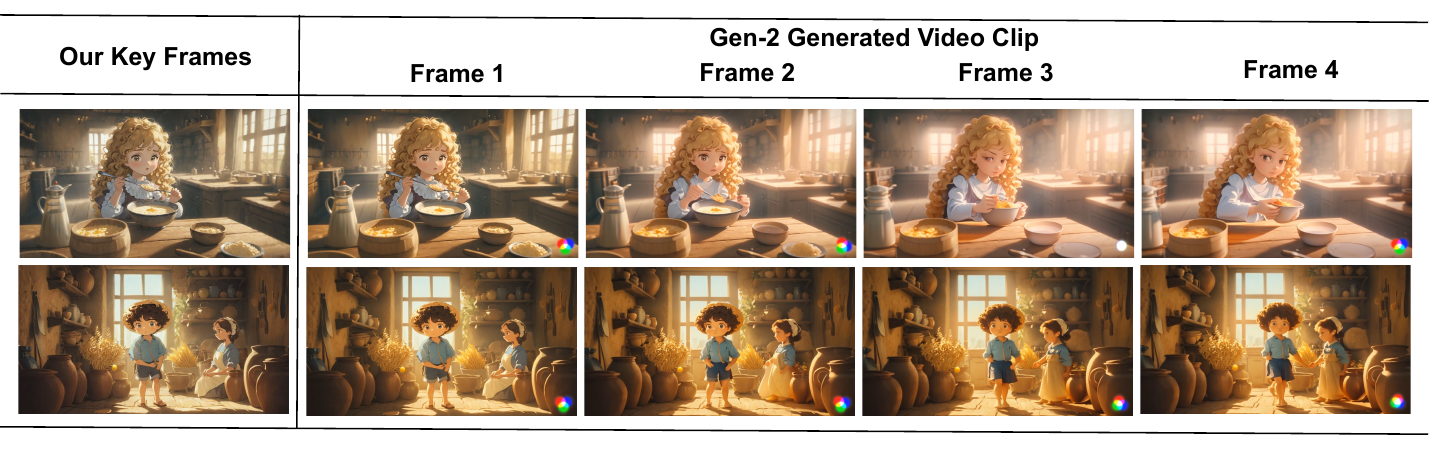}
    \caption{\textbf{Gen-2~\cite{esser2023Structure} generated video clip according to our generated keyframes.} We use a keyframe to generate a 4-second video clip, and show the frames of the video clip.}
    \label{fig:generated_video_clip}
\end{figure}

\subsubsection{The Extension of Utilities} 
Our system could integrate additional units as utilities for enhancing storytelling. Here we employ \revise{an animating unit, i.e., Runway Gen-2~\cite{esser2023Structure},} to transform each generated keyframe into a 4-second video clip, as depicted in Figure~\ref{fig:generated_video_clip}. With Gen-2~\cite{esser2023Structure}, shots such as ``Goldilocks'' consuming porridge and the interaction in ``Epaminondas and Auntie'' are dynamically enriched, contributing to an immersive storytelling experience. So it is feasible for other utilities, \revise{such as Gen-2~\cite{esser2023Structure} and Sora~\cite{Sora}}, to serve as downstream applications of our system.

\subsection{Comparison with Other Methods}

Here we compare our AesopAgent with other methods.  We analyze the user study with ComicAI~\cite{comicai} and compare the storytelling ability with Human Design~\cite{human_design}, NUWA-XL~\cite{DBLP:conf/acl/YinWYWWNYLL0FGW23}, and AutoStory~\cite{DBLP:journals/corr/abs-2311-11243}.

\subsubsection{User Study with ComicAI}

\begin{table}[]
\begin{tabular}{l|ccc|c}
\hline \hline
\textbf{Methods}  & \textbf{Fidelity (50\%)}      & \textbf{Rationality (30\%)}   & \textbf{Element State(20\%)}   & \textbf{Overall}              \\ \hline
ComicAI~\cite{comicai}           & 53.61                & 81.95                & \textbf{93.54}                & 70.10                \\
AesopAgent (Ours) & \textbf{71.56}                & \textbf{86.08}                & 90.67                & \textbf{79.74}      \\ \hline    \hline    
\end{tabular}
\caption{\textbf{The results of the user study on generated image quality.} We collect more than 200 images from 10 stories, and compare their quality on fidelity, rationality, and Element State.}
\label{table:user_study}
\end{table}

We evaluate the image quality produced by two methods, (i.e., AesopAgent and ComicAI~\cite{comicai}). The qualitative results of ComicAI~\cite{comicai} have been shown in Figure~\ref{fig:utility_modules}, here we analyze the user study. Specifically, we randomly select 10 stories to produce above 200 shot images, then conduct manual scoring for each shot image. The evaluation is conducted on three dimensions: fidelity, rationality, and element state. Fidelity measures how closely the elements in the generated images correspond to those in the scripts, with deductions made for missing or misaligned basic elements. Rationality assesses the social realism of depicted elements, deeming elements that are non-existent in reality would be considered irrational (e.g., half-horse). Element State mainly analyzes spatial relationships between visual elements and their environment.

The average scores of these dimensions across the 10 stories are listed in Table~\ref{table:user_study} where these images are given by five annotators. To determine the overall image quality, we employed a weighted sum, assigning weights of 50\%, 30\%, and 20\% to fidelity, rationality, and element state, respectively. Our AesopAgent obtains a slightly lower score on element state since our generated images contain more elements, making it difficult to organize every element precisely. The results also indicate that AesopAgent surpasses ComicAI~\cite{comicai} by 17.95 points in fidelity and by 4.13 points in rationality. This superior performance is credited to our Utility Layer. In detail, the image composition and character description refining module visually represents key script elements; the character inpainting module inpaints the details within these elements; and the image style consistency module ensures the cohesion of the overall style. This structured approach allows our generated images to accurately reflect the script requirements, thereby outperforming ComicAI~\cite{comicai} in the overall score.

\begin{figure}
    \centering
    \includegraphics[width=1\textwidth]{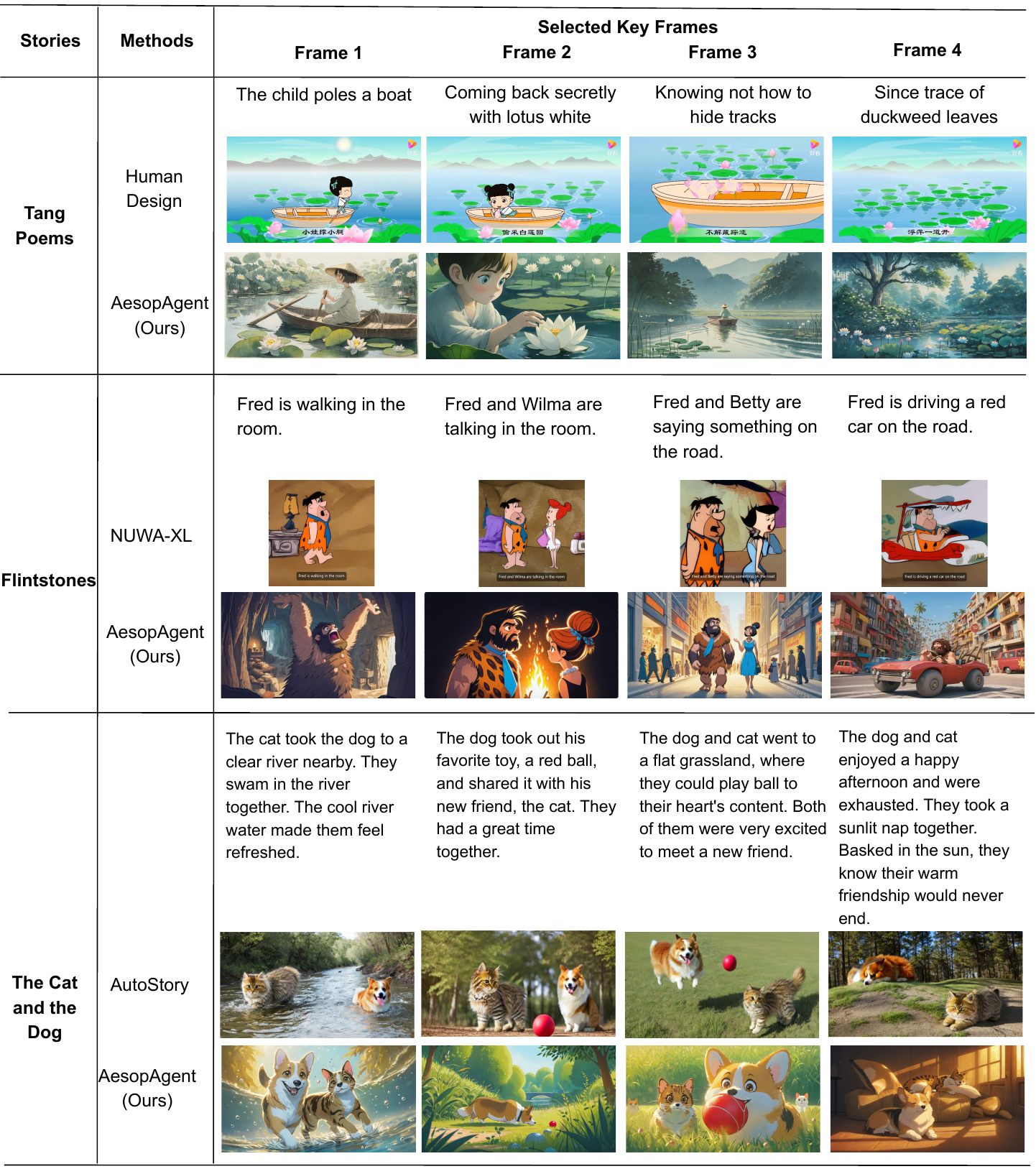}
    \caption{\textbf{Qualitative comparison with other methods.} We compare the keyframes of our AesopAgent with the other three methods (i.e., Human Design~\cite{human_design}, NUWA-XL~\cite{DBLP:conf/acl/YinWYWWNYLL0FGW23}, and AutoStory~\cite{DBLP:journals/corr/abs-2311-11243}) on three stories. The semantic meaning of each frame is listed above the corresponding generated image.}
    \label{fig:comparison_methods}
\end{figure}

\subsubsection{Qualitatively Comparison}

Here we compare our method with other methods. As shown in Figure~\ref{fig:comparison_methods},  our method compares with Human Design~\cite{human_design}, NUWA-XL~\cite{DBLP:conf/acl/YinWYWWNYLL0FGW23}, and AutoStory~\cite{DBLP:journals/corr/abs-2311-11243} in three different stories: ``Tang Poems'', ``Flintstones'', and ``The Cat and the Dog''.
In the ``Tang Poems'', Human Design tends to use simplistic visuals to convey the story's content; while our method would render more detailed and aesthetically pleasing scenes. For the story ``Flintstones'', it is obvious that our method excels at generating the shot with actions (e.g., walking, dialogues involving multiple characters, and driving a car), all of which are more vivid compared with NUWA-XL~\cite{DBLP:conf/acl/YinWYWWNYLL0FGW23}. In the story ``The Cat and the Dog'', our method successfully captures the playfulness and friendship between the cat and dog, whereas AutoStory~\cite{DBLP:journals/corr/abs-2311-11243} simply places the cat and the dog in the same frame without capturing their interaction.

\section{Conclusion}

We propose AesopAgent, an agent system capable of converting user story proposals into videos. This system consists of two primary components: the Horizontal Layer and the Utility Layer. In the Horizontal Layer, we iteratively construct E-RAG and K-RAG based on expert experience and professional knowledge, leading a \textit{RAG-based evolutionary system}. This approach facilitates efficient \textit{task workflow orchestration}, prompt optimization, and utilities usage. The Utility Layer focuses on providing utilities to ensure \textit{consistent image generation} through image composition rationality, consistency across multiple characters, and image style consistency. Furthermore, AesopAgent assembles images, audio, and narration into videos enhanced with special effects.
Experiments demonstrate that it is feasible to use agent techniques to optimize workflow design and utilities usage for complex tasks like video generation. The generated scripts and images of our AesopAgent obtain higher scores than other similar systems, such as ComicAI~\cite{comicai} and Artflow~~\cite{artflow}. And its overall narrative capability surpasses the previous research, including NUWA-XL~\cite{DBLP:conf/acl/YinWYWWNYLL0FGW23} and AutoStory~\cite{DBLP:journals/corr/abs-2311-11243}. \revise{Our system can be further integrated with additional downstream AI utilitise (e.g., Gen-2~\cite{esser2023Structure} and Sora~\cite{Sora}) to meet the requirements of individual users.}
Our future work aims to meet a broader range of user requirements, such as scripts of various themes and genres, a greater diversity of styles, and user-defined characters. This endeavor could even extend to assist the movie generation. At the same time, we will develop a more intelligent agent system, which can better understand the user's intention and complete the user's complex tasks quickly and conveniently.

\bibliographystyle{plain}
\bibliography{egbib}

\end{document}